%% file: acl_latex.tex
\documentclass[11pt]{article}

\usepackage[preprint]{acl}

\usepackage{times}
\usepackage{latexsym}

\usepackage[T1]{fontenc}
\usepackage[utf8]{inputenc}

\usepackage{microtype}

\usepackage{inconsolata}

\usepackage{amssymb} % For checkmark and X symbols  
\usepackage{booktabs} % For formal tables  
\usepackage{xcolor} % For colored text
\usepackage{pifont}
\usepackage{graphicx}
\usepackage{tabularx} 
\usepackage{makecell} 
\usepackage{graphicx} 
\usepackage{multicol}
\usepackage{multirow}
\usepackage{enumitem}
\usepackage{tcolorbox}
\usepackage{color,xcolor,colortbl}
\usepackage{array}
\usepackage{listings}
\usepackage{longtable}
\usepackage{xspace}
\usepackage{wrapfig}
\usepackage{ulem}    % strike
\tcbuselibrary{skins, breakable, theorems}
\usepackage{caption}
\usepackage{stfloats}

\definecolor{Blue}{HTML}{6A9AD0}  
\definecolor{Red}{rgb}{0.746, 0.0039, 0}  
\definecolor{Green}{HTML}{7EAB55}  
\definecolor{Orange}{HTML}{DE8344}
\definecolor{Pink}{HTML}{FF7AA1}
\definecolor{Yellow}{HTML}{FFC000}
\definecolor{deepgreen}{rgb}{0.0, 0.5, 0.0}

\newcommand{\cmark}{{\color{deepgreen}\ding{51}}}  % V 
\newcommand{\xmark}{{\color{red}\ding{55}}}    % X 
\newcommand{\eg}{\normalfont\textit{e.g.}\xspace}
\newcommand{\ie}{\normalfont\textit{i.e.}\xspace}

\newcommand{\method}{\textsc{Thread}\xspace}

\title{\method: A Logic-Based Data Organization Paradigm for How-To Question Answering with Retrieval Augmented Generation}

\author{Kaikai An\textsuperscript{1,2}, 
Fangkai Yang\textsuperscript{2}, 
Liqun Li\textsuperscript{2}, 
Junting Lu\textsuperscript{1,2}, 
Sitao Cheng\textsuperscript{2,3}, \\
\textbf{Shuzheng Si\textsuperscript{1},}
\textbf{Lu Wang\textsuperscript{2},}
\textbf{Pu Zhao\textsuperscript{2},}
\textbf{Lele Cao\textsuperscript{2},}
\textbf{Qingwei Lin\textsuperscript{2},}\\ \textbf{Saravan Rajmohan\textsuperscript{2},} \textbf{Dongmei Zhang\textsuperscript{2},} \textbf{Baobao Chang\textsuperscript{1}}\\
\textsuperscript{1} Peking University 
\textsuperscript{2} Microsoft  
\textsuperscript{3} Nanjing University \\
\texttt{ankaikai@stu.pku.edu.cn,fangkaiyang@microsoft.com,chbb@pku.edu.cn}\\
\href{https://kkk-an.github.io/thread.github.io/}{\textcolor[rgb]{0.80, 0.47, 0.74}{\texttt{thread.github.io}}}
}

\begin{document}
\maketitle

\begin{abstract}
\input{main/abstract}
\end{abstract}

\section{Introduction}
\input{main/introduction}
\section{Methodology}
\input{main/method}
\section{Experiments}
\input{main/experiment}
\section{Analysis}
\input{main/analysis}

\section{Related Work}
\input{main/relatedwork}
\section{Conclusion}
\input{main/conslusion}

\section*{Limitations}
\input{main/limitations}

\normalem
\bibliography{custom}

\newpage
\appendix
\input{main/appendix}

\end{document}

%% file: main/abstract.tex
Recent advances in retrieval-augmented generation (RAG) have substantially improved question-answering systems, particularly for factoid `5Ws' questions. However, significant challenges remain when addressing `1H' questions, specifically how-to questions, which are integral for decision-making and require dynamic, step-by-step responses. 
The key limitation lies in the prevalent data organization paradigm, chunk, which commonly divides documents into fixed-size segments, and disrupts the logical coherence and connections within the context. To address this, we propose \method, a novel data organization paradigm enabling systems to handle how-to questions more effectively. Specifically, we introduce a new knowledge granularity, `logic unit' (LU), where large language models transform documents into more structured and loosely interconnected LUs.
Extensive experiments across both open-domain and industrial settings show that \method outperforms existing paradigms significantly, improving the success rate of handling how-to questions by 21\% to 33\%. 
Additionally, \method demonstrates high adaptability across diverse document formats, reducing retrieval information by up to 75\% compared to chunk, and also shows better generalizability to `5Ws' questions, such as multi-hop questions, outperforming other paradigms.

%% file: main/introduction.tex
Question answering (QA) is a foundational research topic in human-machine interaction~\citep{allam2012question}.
Among the most advanced techniques, retrieval-augmented generation (RAG) enhances QA systems by organizing external documents into fixed-size chunks and retrieving relevant knowledge~\citep{shao-etal-2023-enhancing,trivedi-etal-2023-interleaving,jiang-etal-2023-active,asai2023self}. 
This routine is particularly effective in handling the `5Ws' questions, such as \textit{`When is Shakespeare's birthday?'}, which typically require the data organization paradigm providing chunks containing the relevant knowledge, \eg triples or documents about the topic entity~\citep{yang-etal-2018-hotpotqa,jiang2019freebaseqa,10.1162/tacl_a_00276,stelmakh2022asqa}.
However, the `1H' questions, derived from Aristotle's Nicomachean Ethics (`how-to' questions,~\citealt{crisp2014aristotle}), remain largely underexplored\footnote{The `5Ws' represent What, Why, When, Where, and Who, and the `1H' stands for How.}
These questions are in high demand in practical applications such as teaching us how to write code to achieve specific goals.

\begin{figure*}[t]
    \centering
    \includegraphics[width=0.92\linewidth]{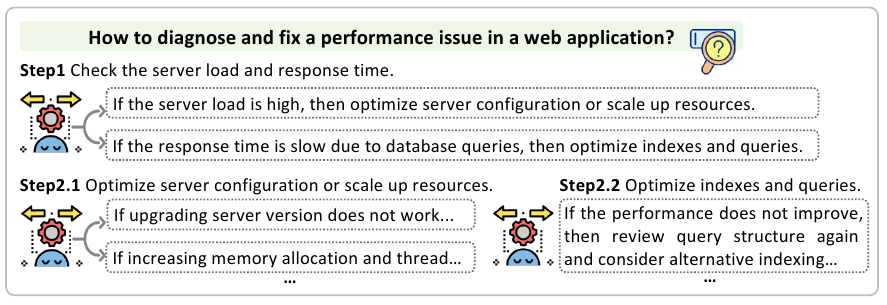}
    \caption{An example of how-to questions with its decision-making process. We omit details such as the actions to check the server load and response time due to limited space.}
    \label{fig:howtoquestions}
    \vspace{-3mm}
\end{figure*}

Central to problem-solving~\citep{polya2014solve} and human learning~\citep{learn2000brain}, how-to questions inherently involve complex processes that require interpretation and analysis~\citep{deng2023nonfactoid}. 
For example, answering the question in Figure~\ref{fig:howtoquestions} `\textit{How to diagnose and fix a performance issue in a web application?}' involves a step-by-step decision-making process, \ie, first checking server load and response time, followed by optimizing server configuration based on user feedback. This dynamic, stepwise nature necessitates RAG systems to guide users through each step, adapting to specific contexts and providing precise and logical information.
However, prevalent chunk-based data organization paradigm\footnote{The term `chunk' here refers to a general document splitting paradigm including chunks, sentences, phrases, etc.}~\citep{Langchain,chen2023dense,gao2023retrieval}, which divides documents into fixed segments, disrupts the logical coherence of content. 
As a result, RAG systems~\citep{asai2023self, shao-etal-2023-enhancing} struggle with how-to questions, often generating excessive, fragmented information that fails to maintain continuity between steps. 
To address this, a paradigm shift is needed, one that preserves the logical structure and stepwise nature of how-to questions.

In this paper, we propose \method, a new logic-based data organization paradigm designed to handle how-to questions. The name \method evokes the idea of `\textit{Pulling on the \textbf{thread}, the whole mystery started to unravel like a sweater.}'~\citep{garcia2011beautiful}. Specifically, we introduce a new knowledge granularity named `logic unit', comprising five key components and four different types (see \S\ref{sec:Definitions} and \S\ref{sec:LUtype}). We employ a two-stage process (depicted in \S\ref{sec:Construction}) to extract logic units (LUs) from documents. The first, optional stage is reformulating the original documents depending on their format and style, and the second focuses on extracting and merging LUs. In this way, \method captures connections within the documents, breaking them into more structured and loosely interconnected logic units. 
When answering how-to questions, the system integrated with \method enables a dynamic interaction manner. First, it retrieves relevant LUs based on their indexed headers. Then, the body of the selected LU provides the necessary content to generate responses for the current step. With user feedback, the linker in LU dynamically connects to other LUs, allowing the system to adapt its responses until the how-to question is comprehensively addressed.

We evaluate the effectiveness of \method through experiments in two open-domain, Web Navigation~\citep{deng2024mind2web}, Wikipedia Instructions~\citep{koupaee2018wikihow}, and one industrial setting, Incident Mitigation~\citep{shetty2022autotsg}. 
Experimental results demonstrate that while existing paradigms struggle with how-to questions, \method excels at handling consecutive steps and consistently outperforms them, particularly in real-world incident mitigation scenario, with success rate improvements ranging from 21.05\% to 33.33\%. Additionally, \method shows great superiority in processing diverse document formats, reducing both the number of retrieval units and the token length required for generation. Finally, we further validate the generalizability of \method on open-domain tasks, where it outperforms the chunk-based paradigm in handling multi-hop questions. 
The main contributions of this paper include:

\vspace{-3mm}
\begin{itemize}[leftmargin=*]
    \setlength\itemsep{0mm}  % Adjusts the space between items
    \item We highlight the challenges faced by current RAG systems in addressing how-to questions. To address the limitation of chunk-based paradigm, we propose \method, a novel data organization paradigm that transforms original documents into structured, interconnected logic units. 
    \item Integrated with \method, our system follows a dynamic interaction manner, guiding users incrementally through each step and adapting to their specific circumstances. Our system also brings more possibilities for an automation pipeline, solving how-to questions more efficiently.
    \item Experimental results demonstrate that \method significantly outperforms existing data organization paradigms across three scenarios. Furthermore, \method efficiently handles various document formats, reducing the retrieval burden, and generalizes effectively to open-domain tasks.
\end{itemize}

%% file: main/method.tex
In this section, \method leverages the internal logic and coherence of documents to construct the knowledge base, making it especially effective for addressing how-to questions. When converting documents into logic units, we classify how-to questions into two types: \textit{\textbf{linear}} and \textit{\textbf{dynamic}}. Linear how-to questions involve a fixed sequence of steps that do not require feedback or decision-making based on intermediate results. In contrast, dynamic how-to questions require decision-making at each step, with the process adapting dynamically based on previous actions, such as the example in Figure~\ref{fig:howtoquestions}.

Below we first introduce `logic unit' (LU) with its components and types, then explain how to extract LUs and construct the \method knowledge base, followed by an illustration of how \method integrates with existing systems. 

\subsection{Logic Unit: Retrieval Unit of Thread}
\label{sec:Definitions}

Unlike the chunk-based paradigm, the logic unit consists of specially designed components that maintain the internal logic and coherence of a document, particularly when bridging consecutive steps in how-to questions. LUs serve as retrieval units that replace traditional chunks in RAG systems, ensuring a more structured and contextually relevant approach to information retrieval. Examples of LU in Table~\ref{tab:example_lu} further illustrate such structure.

\textbf{\textit{Prerequisite.}}
  The prerequisite component acts as an \textit{information supplement}, providing the necessary context to understand the LU. For example, an LU may include domain-specific terminology such as entities or abbreviations. 
  The prerequisite explains these terms and can generate new queries to retrieve LUs with more detailed information. Without this context, passing these LUs to an LLM-based generator could lead to hallucinations. Additionally, the prerequisite can function as an \textit{LU filter}, containing constraints that must be met before the LU is considered in answer generation. This filtering ensures only

\textbf{\textit{Header.}} 
  The header summarizes the LU or describes the intention it aims to address, depending on the type of LU (refer to \S\ref{sec:LUtype}). 
  For example, the header could be the name of a terminology if LU describes a terminology; if the LU describes actions to resolve a problem, the header describes the intent or the problem LU aims to resolve. Different from chunk that indexes the entire content, we use the header for indexing which serves as the key for retrieving the LU based on a query.

\textbf{\textit{Body.}}
  The body contains detailed information on the LU, which is the core content fed into the LLM-based generator to generate answers. It includes specific actions or necessary information such as code blocks, detailed instructions, etc. This detailed content helps resolve the query mentioned in the header or provides a detailed explanation.

\textbf{\textit{Linker.}}
  The linker acts as a bridge between logic units, enabling the dynamic process of how-to questions. Unlike the chunk-based paradigm, which relies on previous retrieval units that often lack direct clues, the linker in \method provides necessary information to generate new queries for subsequent retrieval. 
  Its format varies by LU type, serving as either a query for retrieving other LUs (``If-Then" condition) or an entity relationship. The edge of knowledge graph in traditional factoid questions is a special linker enabling navigation between related entities. When no further LUs are connected, the linker remains empty, isolating the current LU.

\textbf{\textit{Meta Data.}}
  The meta data includes information about the source document from which the LU is extracted, such as the document title, ID, date, and other relevant details. This meta data is crucial for updating LUs when the source documentation is revised and reprocessed.

\subsection{Logic Unit Type}
\label{sec:LUtype}
When converting documents into logic units, \method expands its scope beyond solving how-to questions, as documents often contain more than just solutions to these types of questions. Below are the common LU types identified in our experiments\footnote{These LU types are summarized from our experiments with both industrial and public datasets. Additional LU types are provided in Appendix \ref{appendix_lu_type}, and more may emerge depending on the specific scenario.}, demonstrating \method's versatility in handling a wide range of document types:

\textbf{\textit{Step.}}
  This is the most common type of LU for resolving how-to questions. Each LU body encapsulates detailed actions, including code snippets and resolution instructions. The LU prerequisite specifies the conditions or actions that must be completed prior to executing the current LU, serving a critical role in determining the appropriate entry point for a solution. For instance, when encountering a problem, multiple resolution paths may exist depending on the specific context. The prerequisite functions as a filtering mechanism during the LU selection process, eliminating candidates that do not satisfy the required conditions. Samely, the linker acts as a bridge between steps, directing the flow to subsequent LUs based on the outcome of the current step.

\textbf{\textit{Terminology.}}
  This type provides detailed explanations of domain-specific terminology. For example, terms may share the same name or abbreviation in LU header but convey different meanings in LU body. 
  The prerequisites for terminology LUs describe scenarios where the terminology typically appears and linkers are usually empty unless referencing extended terminology that depends on it.

\begin{figure*}[!t]
    \centering\includegraphics[width=\linewidth]{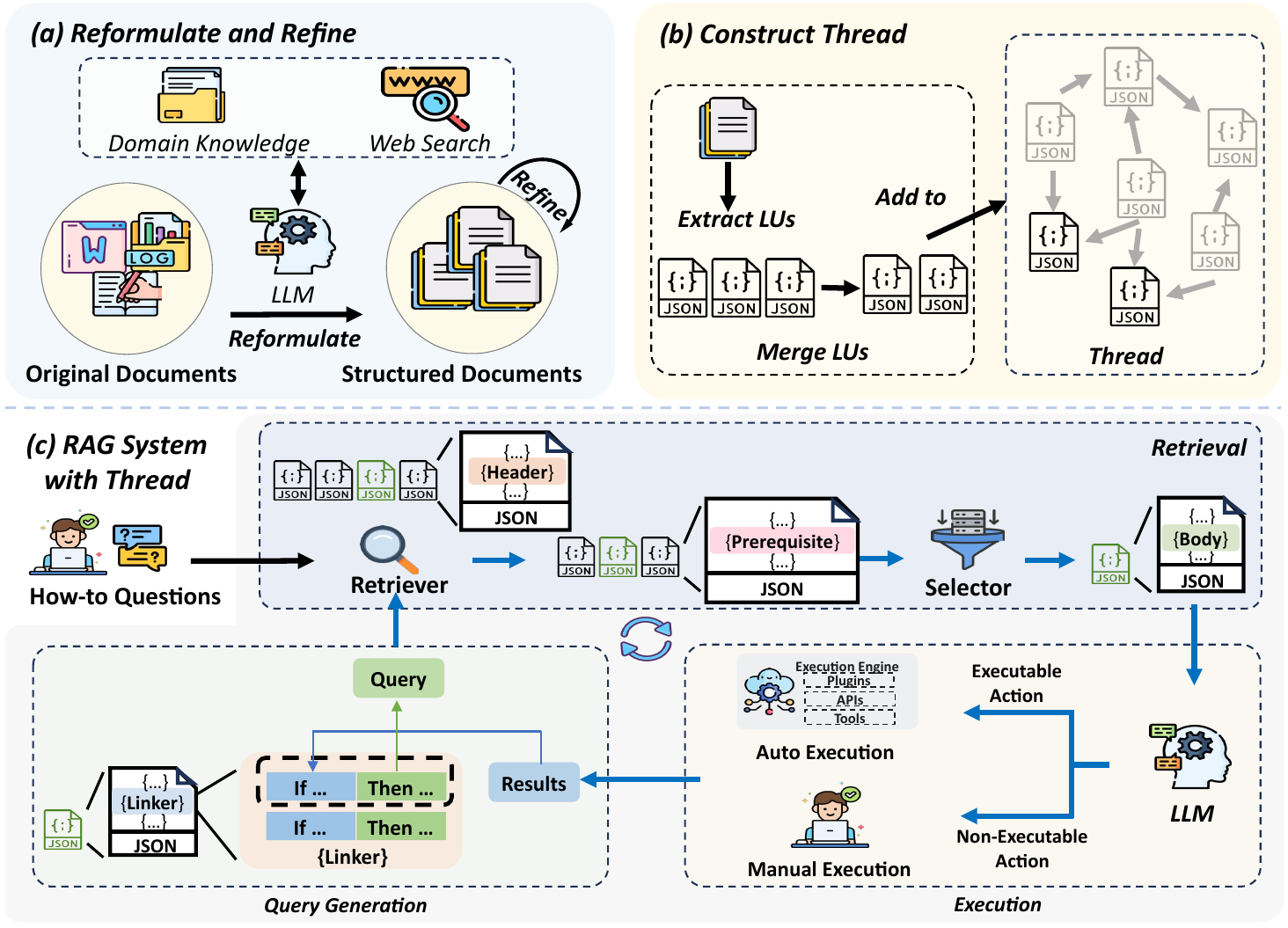}
    \caption{The upper part shows the construction process of \method, a two-stage process including reformulating documents into structured ones (a) and extracting and merging logic units (b). The bottom (c) illustrates the RAG system integrated with \method. It retrieves relevant LUs based on query-\textcolor{orange!90}{Header} similarity and filters out LUs that do not meet the current \textcolor{Pink!120}{Prerequisites}. The selected LUs are passed to LLMs to generate actions based on \textcolor{Green!120}{Body} for execution. After execution, the \textcolor{Blue!120}{Linker} matches the results and generates a new query for the next retrieval iteration. 
    }
    \label{fig:thread}
    \vspace{-5mm}
\end{figure*}

\subsection{Thread: LU-based Knowledge Base}
\label{sec:Construction}

In practice, documentation is often unstructured and varies in format and style. Our approach to converting documentation into \method involves a two-stage process to obtain LUs, as shown in the upper part of Figure \ref{fig:thread}.

\noindent \textbf{Documentation Reformulation.}
This stage is optional, depending on the quality of the documentation. For example, in software engineering, Troubleshooting Guides (TSGs) often vary in format, include diverse types of information, and lack readability and detail, negatively impacting productivity and service health~\citep{shetty2022autotsg}. Due to such format, where some are clearly outlined and others are disordered, we avoid directly extracting LUs from original documents. 

Instead, we first reformulate these documents into structured formats. By leveraging LLMs, we enhance the LLMs' in-domain understanding by providing search capabilities and domain-specific context. This is followed by a refinement step to prevent overlooking details or hallucinating information. Figure~\ref{fig:thread} (a) shows the reformulation stage.
But it is unnecessary for well-written documents like product help docs, which typically follow a linear how-to format. We instantiate this process by example in Appendix \ref{appendix_instantiation}.

\noindent\textbf{LU Extraction and Merge.}
After reformulation, multiple LUs of varying types can be extracted from a single structured document (shown in Figure~\ref{fig:thread} (b)). Unlike chunk-based data organization commonly with fixed chunk sizes, LU granularity depends on content. For example, solutions to linear how-to questions typically form a single path from start to completion, with interconnected steps and no multiple execution outcomes encapsulated in one LU, such as an FAQ LU. However, for dynamic how-to questions with multiple possible outcomes, it is better to have one step per LU (Step LU), with Linkers navigating to the next LUs. Note that in dynamic how-to questions, not every step has multiple execution outcomes. If only one next step exists, the LUs can be merged to include both current and subsequent steps. Additionally, LUs with similar Headers and Bodies should be merged, extending the Prerequisite and Linker. More details about merging are provided in Appendix \ref{tab:lu_merge}.

\noindent\textbf{LU update.}
In industry, documentation is often updated with each product version release. When this happens, we redo the above steps only for the updated documentation, identifying LUs in \method with their Meta Data and replacing outdated LUs.
As we extract and merge LUs, the collection of LUs from all documents forms the knowledge base, which serves as an essential component compatible with the current RAG system.

\begin{table*}[t]
    \small
    \centering
    \begin{tabular}{lccccc|cc}
         \toprule
         \textbf{Dataset} & \textbf{\#Docs}  & \textbf{\#Tasks} & \textbf{\#Steps} & \textbf{\#Chunks} & \textbf{\#LUs} & \textbf{Dynamic} & \textbf{Executable}\\
         \midrule
         Mind2Web & 490 & 252 & 2094 & 6210 & 1089 & \cmark & \cmark\\
         WikiHow & 97 & 97  & 2140 & 4225 & 774  & \xmark & \xmark  \\
         IcM & 56 & 95 & 323 & 413 & 378  & \cmark & \cmark \\
         \bottomrule
    \end{tabular}
    \caption{The statistics and characteristics of datasets, including the number of documents, LUs, etc.}
    \label{tab:data}
\vspace{-5mm}
\end{table*}

\subsection{Integrate Thread with QA System}
\label{sec:qasystem}

To demonstrate how \method works, we use dynamic how-to questions as an example. Figure~\ref{fig:thread} (c) shows the RAG system incorporating our \method data organization paradigm.

LUs are indexed by their Headers. When an initial how-to question is submitted, the Retriever identifies the top-K most relevant LUs based on query-header similarity. The LLM-based Selector then checks the prerequisites of these LUs and filters out those that do not meet the current prerequisites, derived from the initial question or any available chat history. 
After selection, the body of the LUs is fed into the LLM-based Generator to produce an answer. If an execution engine is available, actions can be executed automatically; otherwise, the answer/action is presented to the user for manual execution. Once the action is executed, the Linker matches one of the possible outcomes and generates a new query for the next retrieval round.

Unlike traditional RAG systems, \method-enabled systems can be potentially fully or semi-automated when integrated with execution engines. This integration offers greater automation and flexibility, as updating LUs automatically updates the system, compared to manually designed pipelines.

%% file: main/experiment.tex
\subsection{Scenarios and Datasets}
\label{sec:scenario}

We evaluate \method on two open-domain scenarios: Web Navigation, Wikipedia Instructions, and one industrial setting: Incident Mitigation.

\noindent \textbf{\textit{Web Navigation.}} 
Mind2Web~\citep{deng2024mind2web} is a dataset designed for web agents to perform complex tasks on real-world websites based on language instructions. 
Each task is treated as a multi-choice question. At each step, the input consists of HTML code, an instruction, and a set of choices, while the output is the selected choice, operation, and an optional value. We treat each one as a `dynamic how-to question', with multiple possible outcomes depending on executed actions.

\noindent \textit{\textbf{Wikipedia Instructions.}}
WikiHow~\citep{koupaee2018wikihow}\footnote{\url{https://www.wikihow.com}} provides step-by-step procedural guides, each titled `How to', with a fixed sequence of steps. This dataset is used to evaluate \method on linear how-to questions.

\noindent \textit{\textbf{Incident Mitigation.}}
IcM~\citep{shetty2022autotsg,an2024nissist}  is critical for managing large-scale cloud services, where engineers rely on Troubleshooting Guides (TSGs) to resolve incidents. Each step can lead to different outcomes based on system states, making this dataset suitable for testing \method on dynamic how-to questions. Unlike the open-domain datasets, we conduct a human evaluation with twenty on-call engineers (OCEs) responsible for incident mitigation. Each OCE mitigates five incidents, with baseline methods randomly assigned per incident to prevent familiarity bias. Our RAG system attempts automated mitigation, and if a failure occurs, an OCE intervenes before the system resumes\footnote{Varying across new-hire and experienced OCEs. Note that one OCE's data was contaminated during the experiment, so we removed that OCE's data.}.

We collect documents from open sources, enterprises, and LLMs across scenarios. Appendix~\ref{appendix_documentforretrieval} outlines the collection process, Appendix~\ref{appendix_example} provides dataset examples, and Table~\ref{tab:data} summarizes dataset statistics.

\subsection{Baselines and Metrics}
Previous work on the Mind2Web dataset, such as MINDACT~\citep{deng2024mind2web}, has not approached it as how-to questions, instead using In-Context Learning (ICL) or Supervised Learning (SL). For fair comparison with our LLM endpoints,\footnote{We use GPT-3.5 and GPT-4 (version 1106-preview).} we re-implement MINDACT using the same demonstrations and include chat history as context. 
In this work, we treat Mind2Web as dynamic how-to questions and address them with our RAG system (Appendix~\ref{appendix_rag} shows how we adapt Mind2Web into how-to questions.). We compare \method against document-based and chunk-based data organization paradigms, also used as baselines for the WikiHow and IcM datasets. Details on evaluation metrics, experimental setup (prompts, etc.), and data organization paradigms are provided in Appendix~\ref{appendix_metrics}, \ref{appendix_exp}, and~\ref{appendix_paradigms}, respectively.

\begin{table*}[!t]
    \centering
    \small
    {\renewcommand{\arraystretch}{0.95}
    \setlength{\tabcolsep}{5pt}
    % \resizebox{0.9\textwidth}{!}{
    \begin{tabular}{lcccccc}
    \toprule
     & & &  \multicolumn{4}{c}{\textbf{Mind2Web Cross-Task}} \\
     \cmidrule{4-7}
    \multirow{-2}{*}{\textbf{Method}} & \multirow{-2}{*}{\textbf{Model}} & \multirow{-2}{*}{\textbf{Paradigm}} & \textbf{Ele. Acc} & \textbf{Op. F1} & \textbf{Step SR} & \textbf{SR} \\
    \midrule
     & w/ GPT-3.5 & ICL & 40.69 & 49.66 & 33.91 & 1.59 \\
    & w/GPT-4 & ICL  & 62.80 & 60.37 & 51.81 & 10.32 \\
    \multirow{-3}{*}{MINDACT \citeyearpar{deng2024mind2web}} & \hspace{1em}w/ Flan-T5\textsubscript{XL}* & SL & 55.10 & \textbf{75.70} & 52.00 & 5.20 \\
    \midrule
    & w/ GPT-4 & Chunk  & 64.23 & 65.96 & 58.45 & 8.73 \\
    & w/ GPT-4 & Doc  & 63.80 & 65.89 & 58.36 & 11.51 \\  
    \rowcolor{blue!10}   \cellcolor{white}\multirow{-3}{*}{RAG}  &  w/ GPT-4 & \method  & \textbf{68.29} & 69.53 & \textbf{61.94} & \textbf{12.30} \\
    \bottomrule
    \end{tabular}
    % }
    }
    \caption{Experimental results on Mind2Web. `*' represents taking results from the original paper. \citet{deng2024mind2web} formulate web navigation as a series of multiple-choice steps, where each step requires selecting the correct HTML code and actions. SR refers to the overall Success Rate.}
    \label{tab:mind2web}
    \vspace{-5mm}
\end{table*}

\subsection{Main Results}
\paragraph{Web Navigation}

Table~\ref{tab:mind2web} presents the overall performance on Mind2Web, comparing our method with baselines and both doc-based and chunk-based RAG methods. We find that incorporating informative documents, regardless of the data organization paradigm, significantly improves performance. RAG methods outperform both ICL and SL approaches, with the doc-based RAG method achieving results comparable to MINDACT’s best performance. Notably, \method surpasses all RAG baselines, improving Ele. Acc by 4.06\%, Step SR by 3.49\%, and SR by 3.57\%. MINDACT-SL achieves the highest Op. F1 due to label imbalance\footnote{Predicting all operations as ``CLICK'' results in an Op. F1 of 79.90\%.}, biasing the model toward frequent operations.

\begin{table}[t]
    \centering
    \small
    \resizebox{\linewidth}{!}{
        \begin{tabular}{lccccc}
            \toprule
            & \multicolumn{5}{c}{\textbf{Incident Mitigation}} \\
            \cmidrule(r){2-6}
            \multirow{-2}{*}{\textbf{Paradigm}} & \textbf{SR} & \textbf{Step SR}& \textbf{P.F. Step SR} & \textbf{HI} &  \textbf{Turns} \\ 
            \midrule
            \rowcolor{green!5} \cellcolor{white} & 40.51 & 60.90 & 60.90 & 30.10 & 3.14 \\
            \rowcolor{green!10} \cellcolor{white}  \multirow{-2}{*}{Chunk}  & 28.95 & 53.16 & 43.05 & 46.84 & 6.84  \\
            \midrule
            \rowcolor{green!5} \cellcolor{white}  & 43.86 & 63.90 & 63.90 & 36.09 & 2.98   \\
            \rowcolor{green!10} \cellcolor{white}  \multirow{-2}{*}{Doc}  &  31.58 & 57.89 & 42.11 & 42.11 & 6.53   \\
            \midrule
            \rowcolor{blue!5}  \cellcolor{white} & \textbf{77.19} & \textbf{88.72} & \textbf{84.21} & \textbf{11.28} & \textbf{2.56} \\
            \rowcolor{blue!10}   \cellcolor{white}  \multirow{-2}{*}{\method} &  \textbf{52.63} & \textbf{84.21} & \textbf{68.95} & \textbf{15.79} & \textbf{5.74}  \\
            \bottomrule
        \end{tabular}
    }
    \caption{Experimental results on Incident Mitigation. Incidents are divided into \textcolor{blue!20}{simple} and \textcolor{blue!30}{hard} groups. P.F. Step SR: the perctenage of successful steps before the first failure. HI: the steps requiring human intervention.}
    \label{tab:incident}
    \vspace{-6mm}
\end{table}

\paragraph{Incident Mitigation}

Table~\ref{tab:incident} highlights the advantages of \method in handling complex dynamic how-to questions. Both chunk-based and doc-based RAG methods struggle with incident mitigation, as their low SR and P.F. Step SR scores indicate difficulty in connecting subsequent steps based on the current state. In contrast, \method achieves the highest performance across all metrics, notably improving SR from 21.02\% to 33.33\%. More importantly, its superior \textit{P.F. Step SR} demonstrates its ability to dynamically link steps based on user feedback, reducing human intervention. As a result, \method not only achieves the highest Step SR but also minimizes interaction turns, effectively mitigating both simple and complex incidents.

\paragraph{Wikipedia Instructions}

Table~\ref{tab:wikihow} compares different data organization paradigms on WikiHow under two settings. For single-turn, where the RAG system generates the entire plan in one step, \method outperforms the doc-based by 5.15\%, providing more concise and effective information. In contrast, the chunk-based achieves a significantly lower SR of 19.59\%, highlighting the challenge of retrieving relevant chunks without maintaining the document's logical flow. For multi-turn, iterative retrieval improves performance over single-turn, demonstrating the benefits of a step-by-step approach for how-to questions. The SR increases from 58.76\% to 68.04\% for the doc-based and from 63.91\% to 72.16\% for \method. As in single-turn case, the chunk-based method disrupts internal logic, resulting in a low SR of 20.62\%. Overall, \method achieves the highest SR of 72.16\%, emphasizing its effectiveness in preserving and modeling step dependencies.
We also extend our experiments to LLaMA3-70B to show the generalibity of \method in Appendix \ref{sec:backbone}.

\begin{table}[t]
    \small
    \centering
    \renewcommand{\arraystretch}{1}
    \setlength{\tabcolsep}{9pt}
    \resizebox{\linewidth}{!}{
        \begin{tabular}{lcccc}
        \toprule
         & \multicolumn{4}{c}{\textbf{WikiHow}} \\
        \cmidrule(r){2-5}
         \multirow{-2}{*}{\textbf{Paradigm}} & \textbf{SR} & \textbf{Precision} & \textbf{Recall} & \textbf{F1} \\ 
        \midrule
         \rowcolor{pink!20}  \multicolumn{5}{c}{\textbf{Single-Turn}} \\
         Chunk & 19.59 & 60.20 & 25.16 & 35.49\\
         Doc & 58.76 & 77.71 & 57.93 & 66.37\\
         \rowcolor{blue!10} \method & \textbf{63.91} & \textbf{83.43} & \textbf{71.48} & \textbf{76.99}\\ 
         \midrule
         \rowcolor{pink!20} \multicolumn{5}{c}{\textbf{Multi-Turn}} \\
         Chunk & 20.62 & 52.95 & 25.54 & 34.45 \\
         Doc & 68.04 & 87.10 & 70.65 & 78.02\\
        \rowcolor{blue!10}  \method & \textbf{72.16} & \textbf{89.77} & \textbf{73.36} & \textbf{80.74} \\
        \bottomrule
        \end{tabular}
    }
    \caption{Experimental results on WikiHow with different interaction manners and paradigms.}
    \label{tab:wikihow}
    \vspace{-3mm}
\end{table}

%% file: main/analysis.tex
\subsection{Ablation on RAG System Settings}
This section presents an ablation study on key components of our RAG system, using the Mind2Web dataset. While retriever and generator variants have been explored in prior work~\citep{gao2023retrieval}, we utilize text-embedding-ada-002~\citep{ada002} for retrieval and GPT-4 as the generator.

\begin{table}[t]
    \centering
    \small
    \renewcommand{\arraystretch}{0.9}
    \setlength{\tabcolsep}{6pt}{
        \begin{tabular}{l|ccc}
        \toprule
        \textbf{Paradigm} & \textbf{Ele. Acc} & \textbf{Op. F1} & \textbf{Step SR} \\
        \midrule

         \rowcolor{green!10} ICL   & \textbf{62.80} & \textbf{60.37 }& \textbf{51.81}  \\
         \hspace{1pt} $w/o.$ historical steps & 56.97 & 59.09 & 50.38 \\
            
        \midrule
         Chunk  & 59.94 & 62.58 & 54.39  \\
         \hspace{1pt} $w/o.$ chunk selection   & \textbf{64.23} & \textbf{65.96} & \textbf{58.45} \\ 
         \rowcolor{green!10} \method   & \textbf{68.29} & \textbf{69.53} & \textbf{61.94} \\
         \hspace{1pt} $w/o.$ LU selection  & 67.05 & 68.43 & 60.79 \\
        \bottomrule
        \end{tabular}
    }
    \caption{Ablation study of integrating chat history and retrieval unit selection on Mind2Web.}
    \label{tab:ablation_mind2web}
    \vspace{-5mm}
\end{table}

\noindent \textbf{\textit{Multi-turn Interaction.}} As shown in Table~\ref{tab:wikihow}, the multi-turn setting outperforms the single-turn setting in answering how-to questions. We adopt the multi-turn setting for all scenarios.

\noindent \textbf{\textit{Chat History.}} Chat history enables the system to reference previous actions and results, improving decision-making. Table~\ref{tab:ablation_mind2web} shows that excluding chat history results in performance degradation. Therefore, we include chat history for all RAG-based methods in our experiments.

\noindent \textbf{\textit{Retrieval Units Selector.}} As described in \S\ref{sec:qasystem}, the Selector identifies the most relevant retrieval units from the top-K retrieved units. Table~\ref{tab:ablation_mind2web} illustrates the effect of the retrieval unit selector. Without the selector, performance drops by 4.29\% in Ele. Acc and 3.38\% in Op. F1. The selector improves all metrics when applied to \method. Unlike LU selection, which filters irrelevant LUs based on prerequisites, chunk selection ignores inter-chunk connections and may exclude relevant chunks. Thus, we activate the selector only for \method.

\subsection{Comparison with Different Paradigms}

We compare several RAG data organization paradigms (details in Appendix \ref{appendix_paradigms}), including Semantic Chunking~\citep{SemanicChunk}, Proposition~\citep{chen2023dense}, recursive chunking~\citep{Langchain} (chunk-based), entire document (doc-based), and GraphRAG~\citep{edge2024local}. As shown in Table~\ref{tab:analysis_mind2web}, \method outperforms all baselines across metrics. While Semantic and Proposition merge semantically similar sentences using LLMs, they fail to capture logical relationships. And GraphRAG underperforms on how-to questions due to its focus on entity-level relations over document logic. In contrast, \method retrieves smaller, logic-driven units, yielding higher efficiency and accuracy. 
Appendix~\ref{case_study} illustrates the difference when handling the question in Table \ref{tab:example_lu} between GraphRAG and \method, with GraphRAG producing flawed outputs. 
Appendix~\ref{appendix_result} analyzes \method’s cost and scalability, confirming its high performance at acceptable costs and potential for large-scale datasets.

\begin{table}[t]
    \centering
    \small
    \centering
    \resizebox{\linewidth}{!}{
    \renewcommand{\arraystretch}{1.15} % 修改行距
    \setlength{\tabcolsep}{1mm} % 设置单元格间距
    \begin{tabular}{l|ccc|c}
    \toprule
    \textbf{Paradigm} & \textbf{Ele. Acc} & \textbf{Op. F1} & \textbf{Step SR} & \textbf{\#Tokens in RU}  \\
    \midrule
    Doc & 63.80 & 65.89 & 58.36 & 663.84 \\  
    Recursive & 64.23 & 65.96 & 58.45 & 695.77 \\   
    Semantic & 65.14 & 67.30 & 59.93 & 1337.16 \\
    Proposition & 62.37 & 64.78 & 56.78 & 790.14 \\
    GraphRAG &   63.20 & 65.09 & 56.90 & 630.75 \\
    \midrule
    \rowcolor{blue!10} \method\textsubscript{w/o.} & 67.05 & 68.43 & 60.79 & 772.67\\
    \rowcolor{blue!10} \method & \textbf{68.29} & \textbf{69.53} & \textbf{61.94} & \textbf{157.10} \\
    \bottomrule
    \end{tabular}
    }
    \caption{Analysis of data organization paradigms.} 
    \label{tab:analysis_mind2web}
    \vspace{-5mm}
\end{table}

\subsection{Superiority over Different Doc Formats}
As real-world documents vary in format, we test our LU extraction method on diverse structures, including structured markdown, hierarchical guidelines, tabular checklists, and narrative documents (details in Appendix~\ref{appendix_mind2web}). Table~\ref{tab:format_mind2web} demonstrates that \method consistently outperforms the chunk-based paradigm across all metrics, improving Ele. Acc by up to 9.61\%, Op. F1 by up to 8.43\%, and Step SR by up to 8.51\%. These results underscore \method's ability to effectively handle a variety of real-world document resources. Notably, \method achieves the highest performance with structured documents, which facilitate the creation of a higher-quality knowledge base.

\begin{table}[t]
    \centering
    \small
    \resizebox{\linewidth}{!}{
    \renewcommand{\arraystretch}{1} % 调整行距
    \setlength{\tabcolsep}{5pt} % 调整列间距
    \begin{tabular}{l|c|cccc}
    \toprule
    \textbf{Format} & \textbf{Paradigm} & \textbf{Ele. Acc} & \textbf{Op. F1} & \textbf{Step SR} \\
     \midrule
     & Chunk  & 64.23 & 65.96 & 58.45 \\
     \rowcolor{blue!10} \cellcolor{white}\multirow{-2}{*}{Structured} & \method & \textbf{68.29} & \textbf{69.53} & \textbf{61.94} \\
     \midrule
     
     & Chunk & 60.60 & 63.46 & 55.06 \\
     \rowcolor{blue!10} \cellcolor{white}\multirow{-2}{*}{Hierarchical} & \method & 66.57 & 67.89 & 60.08 \\
     \midrule
     
     & Chunk & 56.30 & 59.26 & 51.43  \\
     \rowcolor{blue!10} \cellcolor{white}\multirow{-2}{*}{Tabular} & \method & 65.71 & 67.69 & 59.55 \\

     \midrule
     & Chunk & 56.63 & 60.39 & 51.66 \\
     \rowcolor{blue!10} \cellcolor{white}\multirow{-2}{*}{Narrative} & \method & 66.24 & 68.22 & 60.17 \\ 
    \bottomrule
    \end{tabular}
    }
    \caption{Analysis of different document formats.}
    \label{tab:format_mind2web}
    \vspace{-5mm}
\end{table}

\begin{figure}[t]
    \centering
    \includegraphics[width=0.95\linewidth]{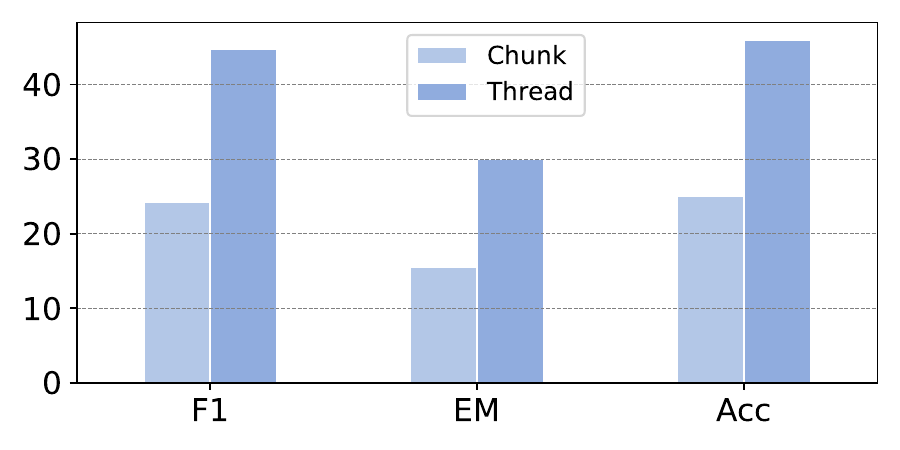}
    \caption{Experimental results on the generalization of \method to 2WikiMultiHopQA.}
    \label{fig:2WikiMultiHopQA}
    \vspace{-5mm}
\end{figure}

\subsection{Generalization to Open Domain Questions}
To explore whether \method can be extended to open-domain questions, we evaluate \method on multi-hop questions, 2WikiMultiHopQA \citep{ho2020constructing}. Specifically, we sample 200 questions of two types, `inference' and `compositional' type, like ``Who is the maternal grandfather of Abraham Lincoln?'' and ``Who is the founder of the company that distributed La La Land film?'' In our experiment, we use the relations between entities as `Linker'. The findings from Figure \ref{fig:2WikiMultiHopQA} demonstrate that THRED largely surpasses the chunk-based paradigm on multi-hop questions, validating the generalization of \method on diverse questions.

%% file: main/relatedwork.tex
\paragraph{Data Organization Paradigm in RAG}
The data organization process is a critical pre-stage of RAG methods where documents are segmented following certain data organization paradigms. The most common data organization paradigm is splitting documents into retrieval units~\citep{gao2023retrieval}. These retrieval units vary in granularity such as phrases, sentences, propositions~\citep{chen2023dense}, chunks~\citep{SemanicChunk}, etc. Coarser-grained units contain more information but introduce redundant noise, while finer-grained units have lower semantic integrity and often require retrieving more units to gather comprehensive information. However, the chunk-based data organization paradigm ignores the logical and relational connections between chunks, potentially disrupting the inherent logic flow in documents. Another paradigm constructs documents into knowledge graphs (KG), where retrieval units include entities, triplets, etc.~\citep{gaur2022iseeq,sen-etal-2023-knowledge,he2024g,wang2024knowledge}.
While these approaches emphasize semantic and lexical similarities between retrieval units, their effectiveness on how-to questions remains limited. This is largely due to the inherent difficulty in capturing the logical structure required by such questions. In particular, current methods such as GraphRAG\cite{edge2024local}, primarily model relationships between different chunks, which fails to account for the more complex, logic-driven connections between chunks that go beyond entity-level similarity.

\paragraph{Information Retrieved by RAG}
The effectiveness of RAG methods depends on the generator's ability to utilize retrieved information and the quality and quantity of that information. Insufficient question-relevant information can cause hallucinations in LLM-based generators~\citep{li2023halueval,zhang2023siren}, making it crucial to improve the retrieval process. Traditional one-round retrieval methods~\citep{guu2020retrieval, NEURIPS2020_6b493230} often fail to gather all necessary information due to their reliance on the similarity between query and retrieval units~\citep{gan2024similarity}.
Advanced RAG methods use query rewriting and expansion~\citep{shao-etal-2023-enhancing,trivedi-etal-2023-interleaving,kim-etal-2023-tree} or iterative retrieval~\citep{shao-etal-2023-enhancing,jiang-etal-2023-active,asai2023self} to collect more information. However, these approaches still struggle with how-to questions, which require making next-step decisions based on the current retrieved units, unless the current retrieved units contain clues that lead to the next step. The main issue is the lack of connections between retrieval units, which prevents effective retrieval and the gathering of sufficient information.

%% file: main/conslusion.tex
In this paper, we address the overlooked category of handling how-to questions in QA systems by proposing \method, a novel data organization paradigm that captures logical connections within documents. By introducing a new knowledge granularity called `logic unit', \method restructures documents into interconnected logic units that are compatible with RAG methods. Extensive experiments show that \method significantly outperforms existing paradigms, improving performance while reducing the knowledge base size and minimizing the information needed for generation.

%% file: main/limitations.tex
This work focuses on evaluating the effectiveness of \method by designing how-to questions in three specific scenarios, covering both linear and dynamic how-to questions. However, several limitations point to future directions. First, as our logic-based knowledge base can coexist with the original chunk-based knowledge base, we only extend our method to multi-hop questions, more open-domain tasks should be considered. 
Next, while extracting logic units involves an initial cost in terms of LLM usage, this is a one-time process. Once the knowledge base is constructed, it provides significant advantages for industrial applications, particularly in terms of subsequent updates and maintenance.

\section*{Acknowledgement}
We thank all reviewers for their great efforts. 
This work is supported by the National Science Foundation of China under Grant No.61876004, which covered personal research efforts, conference registration fees of K.K. An, S.Z. Si and B.B. Chang.

%% file: main/appendix.tex
\section{Scenarios and Datasets}

\subsection{Documents under Each Scenario}
\label{appendix_documentforretrieval}

As the Mind2Web dataset lacks relevant documents, we create retrieval documents tailored to the RAG system.  Assuming that each website has help docs applicable across different tasks, we use the "Cross-Task" test set, selecting examples from the training set to craft informative documents for each website. And we follow \citet{wang2023improving} to brainstorm different formats of documents. More details about collecting documents are shown in Appendix \ref{appendix_mind2web}. 

For the WikiHow dataset, we utilize publicly available Windows Office Support Docs\footnote{\url{https://github.com/MicrosoftDocs/OfficeDocs-Support}} as retrieval documents. We select around 100 tasks from WikiHow tagged with Microsoft products like Word, PowerPoint, and Teams, each containing 10 to 40 steps related to Windows operations.

For the IcM dataset, we collect 56 TSGs from an enterprise-level engineering team responsible for a large-scale cloud platform. The selected incidents in the IcM dataset can be resolved using the knowledge provided in these TSGs.

\subsection{Example of Each Dataset}
\label{appendix_example}
We list one example of each dataset in Table \ref{tab:example_dataset}, and we distinguish how-to questions into \textbf{\textit{`linear'}} and \textbf{\textit{`dynamic'}} types, where linear how-to questions often involve a fixed sequence of steps
that do not require feedback or decision points based on intermediate outcomes.

\begin{table*}[]
    \centering
    \small
    \begin{tabular}{p{0.15\linewidth}|p{0.8\linewidth}}
    \toprule
    Dataset & Example \\ 
    \midrule
    Mind2web \newline (Dynamic) & 
\begin{lstlisting}[basicstyle=\ttfamily\scriptsize, breaklines=true]
<html> ... </html>
Based on the HTML webpage above, try to complete the following task
Task: Book the lowest-priced and quickest flight for 5 adults and 1 child on May 20 from Mumbai to any airport near Washington.
Previous actions:
None
What should be the next action? 
Please select from the following choices (If the correct action is not in the page above, please select A. 'None of the above'):
A. None of the above
B. <div id=0> <input radio triptype roundtrip true /> <label> <span>
C. <label id=1> <span> Search flights one way </span> <span> One
D. <a id=2> <h3> Celebrate World Wish Day </h3> <p> Support
E. <h2 id=3> Help </h2>
F. <a id=4> <img follow us on twitter - /> </a>
C.  Action: CLICK
\end{lstlisting}    \\
    \midrule
    WikiHow \newline (Linear) & 
\begin{lstlisting}[basicstyle=\ttfamily\scriptsize, breaklines=true]
"Problem": "How to Add Captions to Tables in Microsoft Word",
"Solution Steps": [
    "Select the table to which you want to add a caption.",
    "Using your mouse, click and drag over the entire table to select it.",
    "Right-click (or ctrl-click) the table and select Insert Caption.",
    "Enter your caption.",
    "Type the caption for this table into the \"Caption\" field.",
    "Select a caption label.",
    "Customize your caption numbers (optional).",
    "Choose where to place your caption.",
    "Click the \"Position\" drop-down menu, and choose whether to place the caption above or below the table.",
    "Click OK to add your caption to the table.",
    "Format your captions."
]
\end{lstlisting} \\
    \midrule
    IcM \newline (Dynamic) & 
\begin{lstlisting}[basicstyle=\ttfamily\scriptsize, breaklines=true]
How to Investigate Service A-To-Service B Connection?

### Step 1: Check Pull Task Execution From the Cluster
        
The direct impact of connection failure is pull task execution will not work. If Service A can continue to pull from Service B, then the incident can be dismissed as false alarm, the feature owner can investigate further to see why Echo fails. This can be visualized by pull task count over time in the last 8 hours in the following query:
    ```kusto
    <Code Block>
    ```

Disregard the last data point, if the data point is always above zero, then consider the alert as false alarm. If the chart sometimes drops to zero one hour ago and the number is low in general (for instance less than 20), it means the customer traffic in the cluster is low. In this case, observe for a longer period of time. If the data point is zero consistently in the past 30 minutes, then it is a real problem, and please Check if Other Clusters In the Region are Impacted.  Otherwise, continue to observe since Service A is pulling Service B just fine.    
\end{lstlisting}
\\
    \bottomrule
    \end{tabular}
    \caption{Examples of each dataset. For Mind2Web, although the test set has fixed options for each step, there are different execution methods for the same task on each website, so it is essentially dynamic.}
    \label{tab:example_dataset}
\end{table*}

\subsection{Evaluation Metrics under Each Dataset}
\label{appendix_metrics}

% For the Mind2Web dataset, each task is treated as a multi-choice question. At each step, the input consists of HTML code, an instruction, and a set of choices, while the output is the selected choice, operation, and an optional value. 
We adapt the evaluation metrics from~\citep{deng2024mind2web}, which include: \textit{Element Accuracy (Ele. Acc)} to evaluate the chosen HTML element; \textit{Operation F1 (Op. F1)} to calculate the token-level F1 score for predicted operations such as ``CLICK'', ``TYPE IN'', etc.; \textit{Step Success Rate (Step SR)}, where a step is successful if both the selected element and predicted operation are accurate; and \textit{Success Rate (SR)}, where a task is successful only if all steps are successful.

For the WikiHow dataset, which contains ground truth steps, we leverage LLMs to extract ``Action Items'' from each ground truth step and generated step, and we use the following metrics: \textit{Precision} ($P=\frac{\# matched\_items}{\# total\_generated\_items}$); Recall ($R=\frac{\# matched\_items}{\# total\_groundtruth\_items}$); \textit{F1}; and \textit{Success Rate (SR)} to assess if the generated steps can successfully complete the task, using LLMs to evaluate (Appendix~\ref{appendix_wiki} shows the evaluation prompt).

For the IcM dataset, which involves task execution, we perform evaluations with OCEs (refer to \S\ref{sec:scenario}) using five metrics: \textit{Success Rate (SR)} indicating the percentage of incidents mitigated automatically by the system without human intervention; \textit{Step Success Rate (Step SR)} representing the percentage of successful steps out of all task steps; \textit{Pre-Failure Step Success Rate (P.F. Step SR)} representing the percentage of successful steps before the first failure; \textit{Human Intervention (HI)} measuring the percentage of steps requiring human intervention; and \textit{Average Turns (Turns)} to measure the average interaction turns between OCEs and the system during incident mitigation.

\subsection{Remaining LU Types}
\label{appendix_lu_type}
\textbf{\textit{FAQ.}}
  This type provides frequently asked questions, supplementing the knowledge base. These LUs are typically isolated, with the LU body offering solutions through sequential steps that address linear how-to questions not reliant on dynamic states. They save time by avoiding the need for sequentially retrieving LUs for common questions. 

\noindent
\textbf{\textit{Appendix.}}
  This type provides additional information relevant to the scenario of LUs, such as examples, background, lookup tables, etc. These LUs serve as supplementary knowledge for LLMs when generating responses or executable plans.

\subsection{Instantiation of \method}
\label{appendix_instantiation}
In Table \ref{tab:example_lu}, we provide an example of the construction process of \method including the original document, reformulated document, and its corresponding logic unit.

\begin{table*}[]
    \centering
    \small
    % \begin{tabular}{l|c}
    \begin{tabular}{p{0.15\linewidth}|p{0.8\linewidth}}
    \toprule
        \multicolumn{2}{c}{Trouble Shooting Guide: How to Investigate Service A-To-Service B Connection?} \\
        \midrule
         Original  &  \colorbox{gray!20}{\#\#\# Step 0: Determine the Region and Cluster Name}
        
        The region and cluster name can be found in the incident title.
        
        \colorbox{gray!20}{\#\#\# Step 1: Check Pull Task Execution From the Cluster}
        
        The direct impact of connection failure is pull task execution will not work. If Service A can continue to pull from Service B, then the incident can be dismissed as false alarm, the feature owner can investigate further to see why Echo fails. This can be visualized by pull task count over time in the last 8 hours in the following query: ***

        Disregard the last data point, if the data point is always above zero, then consider the alert as false alarm. If the chart sometimes drops to zero one hour ago and the number is low in general (for instance less than 20), it means the customer traffic in the cluster is low. In this case, observe for a longer period of time.
        If the data point is zero consistently in the past 30 minutes, then it is a real problem, and please Check if Other Clusters In the Region are Impacted.  Otherwise, continue to observe since Service A is pulling Service B just fine.
        
        ...
        \\
        \midrule
        Reformulated & \colorbox{gray!20}{\#\# 1.Check Pull Task Execution From the Cluster.}

        \#\#\# Prerequisite

        The region and cluster name can be found in the incident title.
        
        \#\#\# Header
        
        Check Pull Task Execution From the Cluster
        
        \#\#\#  Body

        Run the following query to check pull task execution from the cluster (please use the cluster name from the previous step) ***

        \#\#\# Linker

        - If the data point is always above zero, then consider the alert as false alarm.[MITIGATE] 
        
        - If the chart sometimes drops to zero one hour ago and the number is low in general, it means the customer traffic in the cluster is low. In this case, observe for a longer period of time.[MITIGATE] 
        
        - If the data point is zero consistently in the past 30 minutes, then it is a real problem, and please Check if Other Clusters In the Region are Impacted.[CONTINUE] 
        
        - Otherwise, continue to observe since Service A is pulling Service B just fine.[MITIGATE] 
        
        ...\\
        \midrule
        Logic Unit & 
        \begin{lstlisting}[basicstyle=\ttfamily\scriptsize, breaklines=true]
{
    "#type#": "step",
    "#meta data#": {
        "#title#": "How to Investigate Service A-To-Service B Connection",
        "#id#: "",
        "#date#": ""
    },
    "#prerequisite#": "The region and cluster name are given.",
    "#header#": "Check Pull Task Execution From the Cluster.",
    "#body#": "Run the following query to check pull task execution from the cluster (please use the cluster name from the previous step):***",
    "#linker#": "If the data point is always above zero, then consider the alert as false alarm.[MITIGATE] If the chart sometimes drops to zero one hour ago and the number is low in general, it means the customer traffic in the cluster is low. In this case, observe for a longer period of time.[MITIGATE] If the data point is zero consistently in the past 30 minutes, then it is a real problem, and please Check if Other Clusters In the Region are Impacted.[CONTINUE] Otherwise, continue to observe since Service A is pulling Service B just fine.[MITIGATE]",
    "#default_parameters#": {
        "<TIME>": "",
        "<CLUSTER NAME>": ""
    }
}

...
        \end{lstlisting}        \\
        \bottomrule
    \end{tabular}
    \caption{An example of reformulated TSG and its corresponding Logic Unit of \method.}
    \label{tab:example_lu}
\end{table*}

\newpage

\section{Experimental Details}
\label{appendix_exp}
\subsection{Incident Mitigation}
\label{appendix_incident}

We take the scenario of incident mitigation to show the instructions about how to construct our knowledge base, including document reformulation, code template extraction, and logic unit selection\footnote{The selection strategy is the same for both chunk and LU selections, which leverages LLMs to select the most relevant retrieval element from the retrieved top-K retrieval elements.}.

\begin{tcolorbox}[title = {Instruction that formulates the original unstructured troubleshooting guide into structured one.}, size=title, colframe = white,colbacktitle = black!65!white, breakable]
\noindent

[System]

You are a helpful troubleshooting guide assistant who helps the user formulate the manual unstructured troubleshooting guide <TSG> into a structured one. The <TSG> is in markdown format, with the first level header describing the incident or problem, and the following second level header providing information related to the incident or problem.

Each second-level subsection can be categorized into the following types: Terminology, FAQ, STEP, and Appendix. Your reformulation should strictly comply with the following definition:\\
- Terminology: firstly, it should be the relationship or connection between terminology about the incident, if not, it can be the explanation or concept of the incident. Sometimes it should be extracted and summarized by yourself.\\
- FAQ: frequently asked questions that help to understand the incident.\\
- STEP: the processes to resolve the incident, and you should make sure its completeness. Usually, steps have causal inner connection, the former step will trigger the next step.\\
- Appendix: the supplement of the incident that is not important or labeled by TSG, usually providing additional resources, data, links and so on.\\

1. You need to identify each second-level subsection, including third-level subsection if needed, analyze its content or purpose, and categorize it accordingly. For those belonging to Step, you should capture the inner connections, such as Causality or Temporal relations, and present them in the correct order.

2. Your returned formulated TSG should be in JSON format. Make sure that the keys originate from these categories: Terminology, FAQ, STEP ad Appendix. Each value should be a list of dictionaries. The keys for them are "prerequisite", "header", "body", and "linker". All values within the lists need to align with the original context, with truthful meaning and necessary **code block**.

3. Importantly, the "linker" is used to imply the dual role of providing the action's result and connecting to the next step using the "if-then" sentence format. 
    You should formulate each step's linker to be "If any results are obtained by executing the corresponding action in the previous step, then **the true intent of the following step** provided here". Implicit linkers like "proceed to the next step." or "then the intent of the following step should be taken into consideration." should be avoided.

4. For each "if" condition at every step in the STEP, it is necessary to add a special token behind the "then" condition within the "linker". The options for these tokens are "[CONTINUE]", "[CROSS]", and "[MITIGATE]". 
    - The token "[CONTINUE]" indicates that the actions corresponding to this "if" condition are part of the continuum within the same TSG's STEPs. 
    - The token "[CROSS]" signifies that the subsequent actions require a transition to a different set of steps that are external to the current TSG's STEPs. 
    - The token "[MITIGATE]" implies that the actions following the "if" condition convey that the incident is mitigated, or necessitate communication with on-call engineers or teams. 
    
    The use of this special token is instrumental in verifying the completeness and structural integrity of the STEP section.\\

<TWO EXAMPLES HERE>  \\

[User]

Here is the <TSG> you need to formulate:\\
\textcolor{blue}{\textbraceleft TSG\textbraceright}
\label{tab:prompt_reformulation}
\end{tcolorbox}

\begin{tcolorbox}[title = {Instruction that extracts code template and default parameters from the source code.}, size=title, colframe = white,colbacktitle = black!65!white]

[System]

You are a helpful assistant that extracts the code template and the default parameters from the provided code instance in <CODE>. <CODE> is a code block that contains several parameters. You should replace those parameters with placeholders and output the code template with placeholders and default parameters. \\

<ONE EXAMPLE HERE>\\

Your response should be in the JSON format as below:
\begin{lstlisting}[basicstyle=\ttfamily\scriptsize, breaklines=true]
{
    "#CODE_TEMPLATE#": where you replace the parameters in <CODE> with placeholders,
    "#DEFAULT_PARAMETERS#": where you keep the parameters in <CODE> as default values. 
}    
\end{lstlisting}

[User]

Here is the <CODE> you need to extract:\\
\textcolor{blue}{\textbraceleft CODE\textbraceright}
\label{tab:prompt_code}
\end{tcolorbox}

% \newpage

\begin{tcolorbox}[title = {Instruction that selects the most relevant logic unit based on user query and chat history.}, size=title, colframe = white,colbacktitle = black!65!white, breakable]

[System]

You are a helpful assistant that selects the most relevant element from <LU\_LIST> based on the user's query in <QUERY>  and chat history in <CHAT\_HISTORY>. Please respond with the JSON format.\\
Each element in <LU\_LIST> is in json format and contains the following fields:

\begin{lstlisting}[basicstyle=\ttfamily\scriptsize, breaklines=true]
{
    "#type#": "the type of the element, select from the following types: Terminology, FAQ, Step, and Appendix.",
    "#meta data#": "the description of the troubleshooting guide.",
    "#prerequisite#": "The prerequisite of this step, before taking the current step, the prerequisites should be finished.",
    "#header#": "The information describes the intent of the <INFO>.",
    "#body#": "The action is the content which troubleshoots the incident or explain the #header#. the action may contain code blocks in markdown format, and parameters are replaced with placeholders",
    "#linker#": "the expected output after taking the #action#. It is defined in the following format in markdown: -If **condition**, then **should_do**. It can contain multiple if-then cases.",
    "#default_parameters#": "the default parameters that could fill in placeholders in code blocks in #body#."
}   
\end{lstlisting}

- The elements in <LU\_LIST> contain possible information that can answer the user's query in <QUERY>. However, they may not be all relevant to the query or useful to answer the user's query. You should select the most relevant element from the <LU\_LIST> based on the user's query in <QUERY>. 

- In particular, you should focus on the following fields in the element: \#header\#, \#body\#. Most importantly, the <QUERY> needs to match with the \#intent\# and the \#body\# has to provide actions to reach the goal of the <QUERY>, please ignore the \#linker\# and do not map the <QUERY> with \#linker\#. 

- As you choosing from <LU\_LIST>, you need to check if all the \#prerequisite\# are met in previous history. If the \#prerequisite\# is not finished, then it should not be chosen.

- Try to select only one element from <LU\_LIST>. If it is not possible to select only one element, you can select multiple elements from <LU\_LIST>: 
\begin{lstlisting}[basicstyle=\ttfamily\scriptsize, breaklines=true]
[
    {
        "INDEX": the index of the element in <LU_LIST>.
        "INTENT": the #header# of the element, the index starts from 0. 
        "EXPLANATION": justify why you select this node.
    }
]
\end{lstlisting}
- If there is no element in <LU\_LIST> that can answer the user's query in <QUERY>, you should try to select the most relevant element to the user's query considering that the user might use the wrong terminology: 
\begin{lstlisting}[basicstyle=\ttfamily\scriptsize, breaklines=true]
[
    {
        "INDEX": the index of the element in <LU_LIST>.
        "INTENT": the #header# of the element, the index starts from 0. 
        "REPHRASED_QUERY": the rephrased query that you think the user is asking about.
        "EXPLANATION": justify why you select this node.
    }
]
\end{lstlisting}
- Unless you are confident that there is no element in <LU\_LIST> that is even close to the user's query:
\begin{lstlisting}[basicstyle=\ttfamily\scriptsize, breaklines=true]
{
    "NO_INFO_EXPLANATION": where you give your explanation.
} 
\end{lstlisting}

- Your answer should be in the JSON format in a list after <RESPONSE>.\\

[User]

<LU\_LIST>: \textcolor{blue}{\textbraceleft LU\_LIST\textbraceright}

<QUERY>:    \textcolor{blue}{\textbraceleft QUERY\textbraceright}

<CHAT\_HISTORY>:    \textcolor{blue}{\textbraceleft CHAT\_HISTORY\textbraceright}

\label{tab:prompt_luselect}
\end{tcolorbox}

\subsection{Mind2web}
\label{appendix_mind2web}

We show the details about the document generation instruction we use, the different formats of documents, and the examples we generate.

\begin{tcolorbox}[title = {Instruction that generates specific format of document for Mind2web dataset.}, size=title, colframe = white,colbacktitle = black!65!white]

[System]

You are adept at performing website navigation tasks, and you will be provided with simulation data from Mind2Web, designed for developing and evaluating generalist agents capable of following language instructions to complete complex tasks on any website.\\

The data includes a step-by-step execution process, each step encompassing HTML code, Tasks, Previous Actions, and the Element and Action of this step. 
Note that the Element comes from the HTML code, and if the correct action is not present on the current page, the Element is None, and you should retrieve it from next step's Previous Actions.\\

Now your task is to write a comprehensive and adaptable reference document that outlines the general process for completing tasks like the given task. This document should serve as a guide for others to perform similar tasks on the same website in the future. So it should not be limited but can use this data to be the example, and should be general enough. \\

Please return the complete reference document that adheres to these guidelines.\\

[User]

The format of the documents should be as follows:
\textcolor{blue}{\textbraceleft FORMAT\textbraceright}

The given execution process is as follows:
\textcolor{blue}{\textbraceleft EXAMPLE\textbraceright}
\label{tab:prompt_docgen}
\end{tcolorbox}

We follow \citet{wang2024knowledge} to brainstorm diverse formats of documents for Mind2web dataset used for retrieval, and the results are listed in Table \ref{tab:doc_format}.

\subsubsection{Details about LU Merge}
\label{tab:lu_merge}

\begin{table*}[h]
\small
    \centering
    \begin{tabular}{p{0.1\linewidth}|p{0.8\linewidth}}
    \toprule
        Format & Description \\
        \midrule
        Structured \newline Markdown & - The document must be structured into sections in markdown format.
        
       - It should include a task overview, introduction, process steps, and conclusion. 
       
       - Each step in the process includes detailed explanations for Intent, Prerequisite, HTML Code Reference, Action, Reason, and Result.
       
       - The Prerequisite is to specify any conditions or prior actions that must be met or completed before proceeding with the current step in the process. 
       
       - Ensure that each step is explicitly connected to the next one, and the result is written in the "if-then" schema where the "Intent" of this step is completed, and the outcome "then" is the next step's Intent.
       
       - The HTML Code Reference gives hints of the Action like some '<button>', '<span>', or other elements or attributes. You need to use the given task as an example.
       
       - The Action comes from "Click", "Type", "Hover", "Press Enter".\\
       \midrule
        Hierarchical Guideline  & 
       - A structured text document with numbered steps for each task.
       
       - Each step includes a title, description, the HTML code involved, and the action to be taken.
       
       - Previous actions are referenced where necessary, with hyperlinks to the relevant steps.
       
       - Appendices for HTML code references, glossary of terms, and FAQs.
     \\
    \midrule
    Tabular \newline Checklist & 
   - A printable checklist with each task and subtask, including checkboxes for completion.
   
   - Each checklist item includes a code snippet and the action required.
   
   - A troubleshooting section that lists common problems and their solutions.
   
   - Tips for what to do when the expected element or action is not available.
   
   - References to more detailed instructions or external resources for complex tasks.
    \\
    \midrule
    Narrative \newline Document & 
    An entire description of the execution process without special structures. \\
    \bottomrule
    \end{tabular}
    \caption{The description of different formats of documents on Mind2web.}
    \label{tab:doc_format}
\end{table*}

For LU merge, we first identify the similar logic units by using the SpaCy library to calculate the textual similarity of LU headers. Then we leverage LLM to merge LUs with the following prompt:

\begin{tcolorbox}[title = {Instruction that merges logic units with similar header.}, size=title, colframe = white,colbacktitle = black!65!white,breakable]

[System]

You are tasked with a set of Logic Units that contain information about different steps in web navigation task. Each unit includes components like type, title, header, prerequisite, body, and linker. Some units have similar intents and can be merged to streamline the process and reduce redundancy. \\

Your task is to merge logic units with similar header into a single unit that combines their prerequisite, body and linker in a logical and coherent manner. \\

- Most importantly, as merging, you should concentrate on the linker, you need to unite the linker with similar intent, and carefully compare their "if" conditions. These conditions should now depend on the title specifics, guiding the user to the appropriate next action based on the context of the task. \\
- And for prerequisite, you should synthesize the prerequisites from the individual units, preserving the original logic and ensuring that the merged unit sets the necessary conditions for the subsequent steps.  \\

The purpose of this merge is to create a more efficient set of instructions that can handle multiple scenarios without repeating steps. \\

Here is an example:   \\

Please only return the merged unit in JSON format, keeping the same structure with the input.   \\

[User]

The logic units you need to merge are as follows:
\textcolor{blue}{\textbraceleft units\textbraceright}
\end{tcolorbox}

\subsection{Wikihow}
\label{appendix_wiki}

\begin{tcolorbox}[title = {Instruction that evaluates the generated answer compared with ground truth for Wikihow.}, size=title, colframe = white,colbacktitle = black!65!white, breakable]

[System]

You are a helpful and precise assistant for checking the quality of the answer. We would like to invite you to evaluate the performance of the system in answering a user's question in <Question>. \\

I will give you the answer generated by the system in <Generation> and the ground truth answer in <Ground Truth> respectively. Your evaluation will contain five sub-evaluation tasks:\\

1. Both two answers contain a list of steps. Your task is to extract action items from the provided steps in both answers. The action item is defined as a combination of action and element. Compare the action items to identify similarities. Output the similar action items. Count the count of similar action items.\\

    - Your answer should contain the extracted two action item sets (in the format as a list of strings).\\
    - Your answer should contain a set of similar action items (in the format of a list of strings). Similar action items are those sharing similar intent or achieving similar goals. Each similar action pair in the list should be in the format of "similar action item from action item set1 / similar action item from action item set2"
    - Your answer should contain the count of similar action items.\\
    
2. Can <Generation> completely solve the user's question?\\
    - Your answer should be "Yes" or "No".\\
    - Your answer should contain the reason(s) for your choice. You should not focus on the length of the answer or the details of the answer, but you should focus on whether the steps could solve the user's question and the quality of the steps compared with the ground truth.\\

Your output should be in the following format in JSON:

\begin{lstlisting}[basicstyle=\ttfamily\scriptsize, breaklines=true]
{
    "Subtask1": {
        "Action items in Generation": ["action item 1", "action item 2", ...],
        "Action items in Ground Truth: ["action item 1", "action item 2", ...],
        "Similar action items": ["similar action item 1", "similar action item 2", ...],
        "Count of similar action items": 2
    },
    "Subtask2": {
        "Choice": "Yes" or "No",
        "Reason": "reason for your choice"
    }
}
\end{lstlisting}

[User]

Here is the user's question <Question>: \textcolor{blue}{\textbraceleft Question\textbraceright}\\
The answer from system <Generation> is: \textcolor{blue}{\textbraceleft Generation\textbraceright}\\
The ground truth answer <Ground Truth> is: \textcolor{blue}{\textbraceleft Ground Truth\textbraceright}
\label{tab:prompt_wiki}
\end{tcolorbox}

\subsection{Details about RAG System}
\label{appendix_rag}
We take the scenario of Mind2web to show the instructions we use in our RAG-based QA system.

\begin{tcolorbox}[title = {Instruction that is used for the baselines of RAG system on Mind2web.}, size=title, colframe = white,colbacktitle = black!65!white, breakable]

[System]

You are a helpful assistant who is great at website design, navigation, and executing tasks for the user. Now please proceed with the <CURRENT\_STEP> and make your choice, remember that only based on the helpful document information from <DOC\_CONTEXT> and the previous step chat history between user and assistant in <CHAT\_HISTORY>.\\

Your response should be in the format of "Answer: C. Action: SELECT Value: Pickup". \\
The answer is A, B, C..., the Action comes from [CLICK, TYPE, SELECT] and the Value is not always needed.\\

[User]

<DOC\_CONTEXT>: \textcolor{blue}{\textbraceleft DOC\_CONTEXT\textbraceright}\\
<CHAT\_HISTORY>: \textcolor{blue}{\textbraceleft CHAT\_HISTORY\textbraceright}\\
<CURRENT\_STEP>: \textcolor{blue}{\textbraceleft CURRENT\_STEP\textbraceright}\\
\label{tab:prompt_baseline}
\end{tcolorbox}

\begin{tcolorbox}[title = {Instruction that is used for the RAG system utilizing \method Paradigm on Mind2Web.}, size=title, colframe = white,colbacktitle = black!65!white, breakable]

[System]

You are a helpful assistant who is great at website design, navigation, and executing tasks for the user. Now please proceed with the <CURRENT\_STEP> and make your choice, remember that only based on the helpful structured document information from <LOGIC\_UNIT>, and the previous step chat history between user and assistant in <CHAT\_HISTORY>.\\

Your response should be in the format of JSON:
\begin{lstlisting}[basicstyle=\ttfamily\scriptsize, breaklines=true]
{
    "CHOICE": the choice you make from A B C ...,
    "ACTION": the corresponding action choosing from ['CLICK', 'TYPE', 'SELECT'],
    "VALUE": the corresponding value if needed,
    "INTENT": the intent of the next step, which should be retrieved and judged from the "if" conditions in #output# from <LU> according to the current step and actions and choose the corresponding "then" outcome, do not guess it based on current Task in <CURRENT_STEP> by yourself unless the <LU> is irrelevant to <CURRENT_STEP>,
}
\end{lstlisting}

[User]

<LOGIC\_UNIT>: \textcolor{blue}{\textbraceleft LOGIC\_UNIT\textbraceright}\\
<CHAT\_HISTORY>: \textcolor{blue}{\textbraceleft CHAT\_HISTORY\textbraceright}\\
<CURRENT\_STEP>: \textcolor{blue}{\textbraceleft CURRENT\_STEP\textbraceright}\\
\label{tab:prompt_thread}
\end{tcolorbox}

\newpage
\label{sec:backbone}
\begin{figure*}[t]
    \centering
    \includegraphics[width=0.85\textwidth]{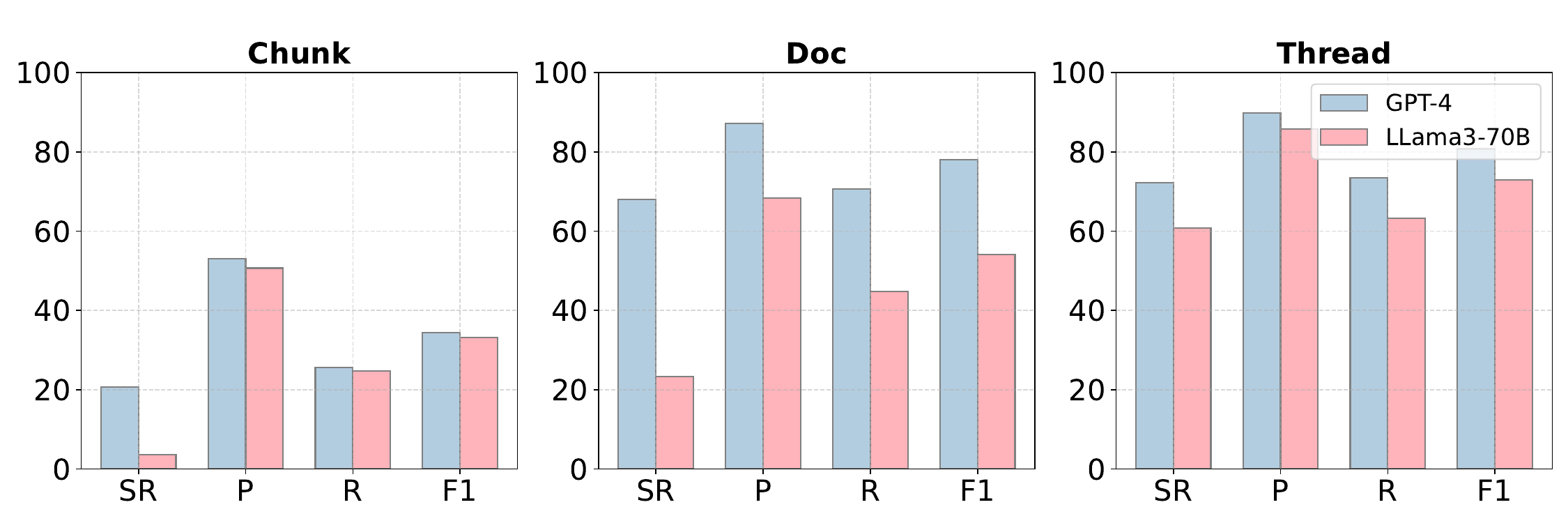}
    \caption{Analysis of using different LLMs on WikiHow (GPT-4 v.s. LLaMA3-70B).}
    \label{fig:llama}
\end{figure*}

\section{Current Data Organization Paradigm}
\label{appendix_paradigms}
From \citet{gao2023retrieval}, current data organization paradigms can be categorized into phrases, sentences, propositions, chunks, and so on.
In our paper, we choose chunks \footnote{We use the implementation by LangChain \url{https://python.langchain.com/v0.2/}} and propositions to compare with our proposed \method\footnote{Note: For chunks, we retrieve the top-5 at each time, and for documents, we only retrieve the top-1.}.

\noindent \textbf{Recursive Chunk.} 
This chunking method splits the original documents using a list of separators, then reassembles them according to specified chunk sizes and overlap sizes. In our experiment, we use different chunk sizes for each dataset: 1000 for Mind2Web, 2000 for IcM, and 300 for WikiHow. The chunk overlap sizes also vary 50 for Mind2Web, 100 for IcM, and 30 for WikiHow.

\noindent \textbf{Entire Document.} 
This method sends the entire document directly into the model, constrained by the document's length and structure.

\noindent \textbf{Semantic Chunk.} \citet{SemanicChunk} proposes splitting chunks based on semantic similarity. The hypothesis is that semantically similar chunks should be grouped together. By comparing the semantic similarity between adjacent sentences, the method identifies ``break points''. If the similarity in the embedding space exceeds a certain threshold, it marks the start of a new semantic chunk.

\noindent \textbf{Agentic Chunk (Proposition).} \citet{chen2023dense}\footnote{For proposition paradigm, we use agentic chunker since the input token of Flan-T5 is limited to 512, \url{https://github.com/FullStackRetrieval-com/RetrievalTutorials}.} introduces the concept of the Proposition Paradigm, which involves extracting independent propositions from original documents. The Agentic Chunk method is based on this paradigm. It first splits the documents into paragraphs, then extracts propositions from each paragraph, and at last merges similar propositions into chunks.

\noindent \textbf{GraphRAG.}
GraphRAG~\citep{edge2024local} is a data organization approach proposed for those query-based summarization, e.g., `what is the theme of this dataset', which constructs KG by extracting entities from chunks and then constructing relationships between them.

\section{Additional Experimental Results}
In all our experiments, we set the temperature of LLMs to 0 and top\_p to 1 for results reproduction.

\subsection{Different LLMs as backbone}

We further conduct experiments on LLaMA3-70B, as shown in Figure \ref{fig:llama}. Although LLaMA3-70B is less powerful than GPT-4, it still demonstrates competitive performance with the help of \method. Results indicate that LLaMA3 struggles with WikiHow questions applying chunk-based or doc-based paradigms. However, with the integration of \method, LLaMA3 not only achieves much better performance in SR, but also narrows the gap with GPT-4. Specifically, LLaMA3 achieves SR and F1 scores of 60.82\% and 72.84\%, respectively, compared to GPT-4's scores of 72.16\% and 80.74\%.
This indicates that while a performance gap remains, particularly with the chunk-based and doc-based paradigms, our \method considerably reduces this disparity between GPT-4 and LLaMA3. These findings highlight the value of our proposed paradigm in enhancing the performance of different LLMs, showcasing its generalizability, robustness, and efficiency in handling how-to questions.

% \newpage
\subsection{Case Study}
\label{case_study}

We compare the outputs from systems equipped with \method and GraphRAG that handles how-to question given in Table \ref{tab:example_lu}, `How to Investigate Service A-To-Service B Connection?'.

\noindent \method can progressively mitigate such questions as follows:
\begin{lstlisting}[basicstyle=\ttfamily\scriptsize, breaklines=true]
Step1: 
When investigating the connection between Service A and Service B, the first step is to check pull task execution from the cluster. Please run the following query to check:
*****

- If the data point is always above zero, then consider the alert as false alarm.
- If the data point is zero consistently, then it is a real problem, and please Check if Other Clusters In the Region are Impacted.

(the outcome of step1: data point is zero consistently)
Step2:
To determine the list of clusters, the quickest way is to check IcM by running
the following query to get the incidents in the last day:
*****
- If the incident count is one, ..
- If the incident count is more than one, ..

Step3:
...
\end{lstlisting}

However, the response generated by GraphRAG is highly flawed. It not only distorts the internal logic within the documents but also retrieves chunks that contain the entities "Service A" and "Service B". After that, it misuses this retrieved information to generate steps that are entirely irrelevant, seemingly getting stuck at start. 
\begin{lstlisting}[basicstyle=\ttfamily\scriptsize, breaklines=true]
To investigate a connection issue between Service A and Service B, you can follow this step:

Verify if Service A is actually making a connection call to Service B. 
You can generally do this by checking the logs or telemetry data where outbound requests are recorded using a Kusto Query:
***

If the connection attempts are not present in the logs, ensure that Service A is configured correctly to make calls to Service B.
If attempts are recorded, you can investigate further based on the `ResultCode` to determine if there are timeouts, rejections, or any other issues that could indicate a problem with the connection.
\end{lstlisting}

\subsection{Cost and Scalability}
\label{appendix_result}
\begin{table*}[t]
    \centering
    \setlength{\tabcolsep}{5pt}
    \resizebox{0.8\textwidth}{!}{
        \begingroup
        % \color{blue}
        \begin{tabular}{lccccc}
        \toprule
        \textbf{Paradigm}       & \makecell[c]{\textbf{Chat Tokens} \\ \textbf{(I/O) / Doc}} & \makecell[c]{\textbf{Embedding } \\ \textbf{Tokens / Doc}} & \makecell[c]{\textbf{Price / Doc (\$)}} & \makecell[c]{\textbf{Chat Model API} \\ \textbf{Call / Doc}} & \makecell[c]{\textbf{Ele. Acc}} \\ 
        \midrule
        Recursive (Chunk) & - & 838 & 0.000084 & - & 64.23 \\
        Semantic & - & 3802 & 0.00038 & - & 65.14 \\
        Proposition & 61836 / 388 & 978 & 0.63 & 28.8 & 62.37 \\
        \method & 2553 / 654 & 654 & 0.045 & 1 & 68.29 \\
        \bottomrule
        \end{tabular}
        \endgroup
    }
    \caption{Cost of different paradigms on Mind2Web.}
    \label{tab:comparison_paradigms}
    \vspace{-4mm}
\end{table*}

\begin{table*}[t]
\centering
{
\renewcommand{\arraystretch}{1}
    \setlength{\tabcolsep}{5pt}
    \resizebox{0.95\textwidth}{!}{
        \begingroup
        % \color{blue}
        \begin{tabular}{lcccccc}
        \toprule
        \textbf{Dataset}       & \makecell[c]{\textbf{Total Tokens}} & \makecell[c]{\textbf{Original} \\ \textbf{Tokens / Doc}} & \makecell[c]{\textbf{Chat Tokens} \\ \textbf{(I/O) / Doc}} & \makecell[c]{\textbf{Embedding} \\ \textbf{Tokens / Doc}} & \makecell[c]{\textbf{Price / Doc (\$)}} & \makecell[c]{\textbf{$\Delta$ Ele. Acc}} \\ 
        \midrule
        Incident Mitigation & 112K & 2002 & 2814 / 1464 & 1464 & 0.072 & 23.68\%-36.68\% ↑ \\
        WikiHow & 200K & 2066 & 4608 / 778 & 778 & 0.069 & 44.3\%-51.54\% ↑ \\
        Mind2Web & 410K & 838 & 2553 / 654 & 654 & 0.045 & 4.06\% ↑ \\ 
        \bottomrule
        \end{tabular}
        \endgroup
    }
}
\caption{Scalability across different scenarios.}
\label{tab:dataset_comparison}
\vspace{-5mm}
\end{table*}

We analyze the preprocessing overhead for constructing the final knowledge base using the Mind2Web dataset and compare \method to other paradigms that also leverage LLMs during preprocessing. From Table \ref{tab:comparison_paradigms}, \method achieves the highest element accuracy (68.29\%) with a balanced trade-off between cost and performance. While its cost is higher than the Chunk and Semantic paradigms, it is significantly lower than the Proposition paradigm, making it suitable for real-world applications. Additionally, in Figure \ref{fig:llama}, we show that Llama3-70B achieves comparable performance with GPT-4, further emphasizing \method's effectiveness and generability.

More importantly, to evaluate \method's scalability, we curate datasets by transforming existing datasets (e.g., Mind2Web) and collecting data from the internet (e.g., WikiHow) and examined its performance on three datasets with different sizes. Table \ref{tab:dataset_comparison} shows that \method consistently outperforms the Chunk paradigm across datasets of varying scales, achieving higher element accuracy while maintaining acceptable costs. This demonstrates \method's scalability in handling both longer documents and larger corpora.